\documentclass[10pt,twocolumn,letterpaper]{article}

\usepackage[pagenumbers]{iccv} % To force page numbers, e.g. for an arXiv version

\usepackage{multirow}
\usepackage{listings}
\usepackage{csquotes}

\definecolor{iccvblue}{rgb}{0.21,0.49,0.74}
\usepackage[pagebackref,breaklinks,colorlinks,allcolors=iccvblue]{hyperref}

\def\ourdata{\textsc{CoPa-SG}}
\def\numrel{86M}

\def\paperID{6350} % *** Enter the Paper ID here
\def\confName{ICCV}
\def\confYear{2025}

\title{\ourdata{}: Dense Scene Graphs with Parametric and Proto-Relations}

\author{Julian Lorenz \quad Mrunmai Phatak \quad Robin Schön \quad Katja Ludwig \\ Nico Hörmann \quad Annemarie Friedrich \quad Rainer Lienhart\\
University of Augsburg\\
Augsburg, Germany\\
{\tt\small \{firstname.lastname\}@uni-a.de}
}

\begin{document}
\maketitle

\begin{abstract}
    % what are scene graphs
    2D scene graphs provide a structural and explainable framework for scene understanding.
    % current work struggles because data is not good enough
    However, current work still struggles with the lack of accurate scene graph data.
    % we use synthetic data to create accurate data
    %We address this issue
    To overcome this data bottleneck, we present {\normalfont \ourdata{}}, a synthetic scene graph dataset with highly precise ground truth and exhaustive relation annotations between all objects.
    Moreover, we introduce parametric and proto-relations, two new fundamental concepts for scene graphs. The former provides a much more fine-grained representation than its traditional counterpart by enriching relations with additional parameters such as angles or distances. The latter encodes hypothetical relations in a scene graph and describes how relations would form if new objects are placed in the scene.
    % show applications of our dataset
    Using {\normalfont \ourdata{}}, we compare the performance of various scene graph generation models.
    We demonstrate how our new relation types can be integrated in downstream applications to enhance planning and reasoning capabilities.
\end{abstract}

%%%%%%%%% BODY TEXT
\section{Introduction}

% Always speak of "traditional" scene graph datasets when referring to existing datasets. Our dataset is new and very different. And awesome. Reviewers must always be remembered of this fact.

% What are scene graphs?
2D scene graphs \cite{first_scenegraph} provide structured representations of scenes, usually depicted in an image,
% Why should we care about scene graphs?
which can be used for further downstream applications like navigation \cite{sg_navigation}, planning \cite{conceptgraphs}, and reasoning \cite{sg_reasoning0,sg_reasoning1}.
Scene graphs represent relations between instances as triples (subject instance, predicate class, object instance), \eg, \textit{(chair, next to,  table)}.
As a structured intermediate step, they contribute to interpretability and robustness.

\begin{figure}
    \centering
    \includegraphics[width=\linewidth]{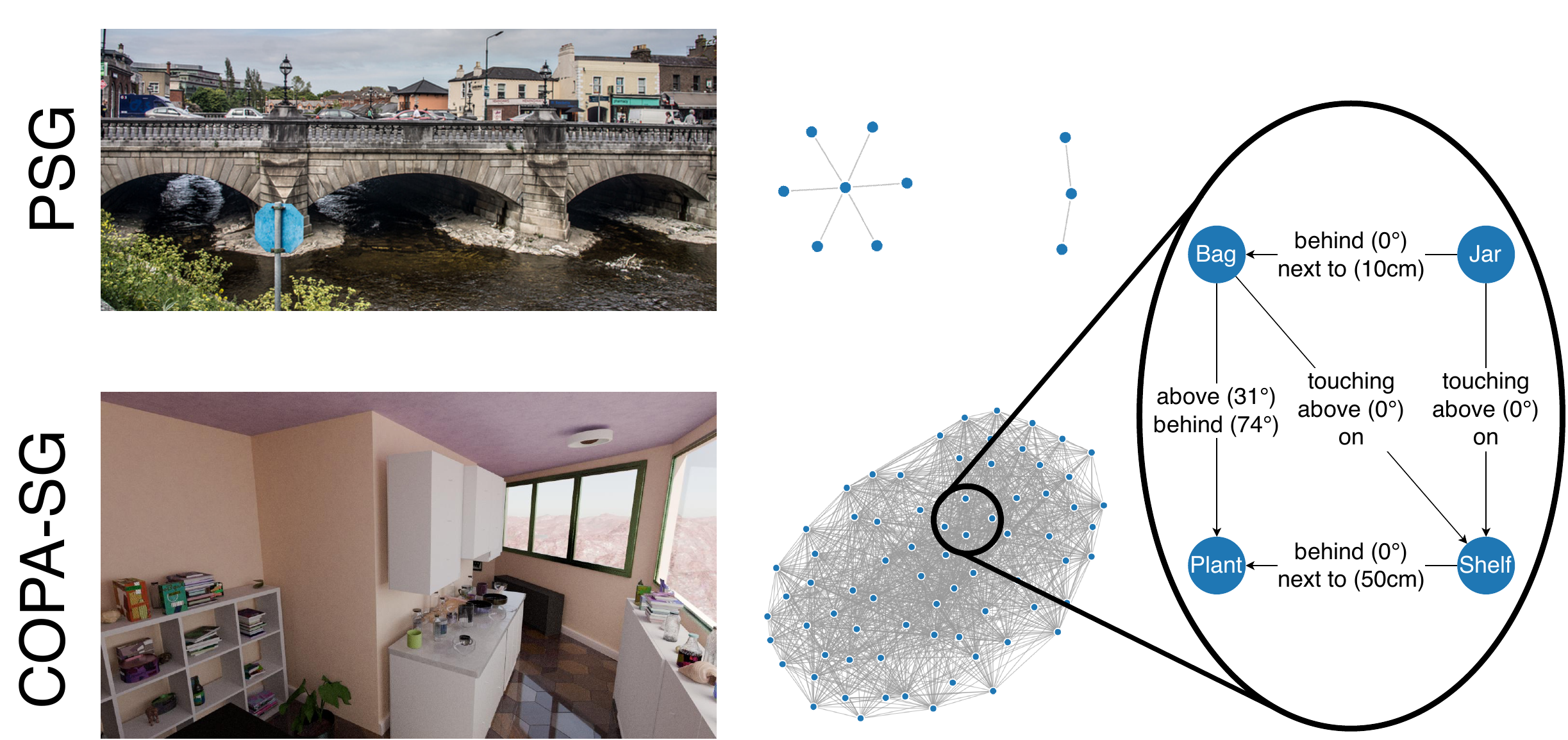}
    \caption{\ourdata{} contains exhaustive scene graph annotations. In contrast, traditional scene graphs focus mostly on salient relations. This makes our dataset more suitable for training and evaluating scene graph generation models that will be employed in downstream applications.
    Note that the magnified graph displays only a subset of the available relations for better readability.}
    \label{fig:example_scene}
\end{figure}

% What is the main problem of traditional scene graph datasets?
Traditional scene graphs are manually annotated based on textual region descriptions of the respective image \cite{visual_genome,psg}, resulting in incomplete and inconsistent ground truths.
% problem 1: incomplete annotations
Existing datasets focus on salient relations, because annotating every single relation in an image by hand is infeasible. Recent work \cite{as1b} resorts to vision-language models for annotation, but these models also focus mostly on salient relations. Thus, existing scene graph datasets are incomplete and miss many valid relations.
% problem 2: inconsistent annotations
The predicate class of a scene graph relation is expressed using a set of prepositions such as \textit{behind} or \textit{left}, which depend on the viewer's location or perspective and may be ambiguous in their spatial interpretation \cite{tyler2003semantics,schneider-etal-2015-hierarchy,schneider-etal-2018-comprehensive}.
Hence, imprecisely defined predicate classes may also contribute to inconsistencies in manual scene graph annotations.
% Hence, another source of bad ground truth quality are imprecisely defined predicate classes.
For some annotators, a chair is behind a table if it is just farther away, for others it must be occluded.
The issue of noise in annotated scene graph datasets has to date mainly been addressed by modifying the training protocol \cite{IETrans,ADTrans,hilo}, but not by directly improving the quality of the benchmarks.

% How do we intend to solve it?
In this work, we present \ourdata{} (\textbf{Co}mplete and \textbf{Pa}rametric \textbf{S}cene \textbf{G}raphs), a new synthetic scene graph dataset with a focus on highly reliable annotations.
Current research demonstrates that including synthetic data into the training pipeline can greatly improve model accuracy \cite{sapiens,depth_anything_v2}. In this paper, we provide a pipeline that can extract \textbf{exhaustive and precise scene graph annotations} for arbitrary synthetic datasets where the underlying 3D scene is available.
% our target applications are robotic agents that use our representation, not humans
In contrast to existing scene graph datasets, \ourdata{} is not biased towards salient relations and provides a complete set of relations for each scene. See \cref{fig:example_scene} for a visual comparison. On average, a scene in \ourdata{} contains more than 72k relations. The relation extraction process is based on a strict set of deterministic rules and does not depend on subjective human interpretation. This reliable annotation method enables the training of scene graph generation models which are better suited for downstream tasks.

% mention parametric relations and proto-relations here
% parametric-relations
To eliminate ambiguous predicate definitions, we introduce the concept of \textbf{parametric relations}. In addition to a traditional predicate label, we store a parameter (\eg an angle or a distance) that enables a more fine-grained representation. We show how existing models can be adapted to the new parametric scene graph generation task.
% proto-relations
Additionally, we introduce \textbf{proto-relations} as a novel technique for representing hypothetical relations.
Given an anchor object and a predicate, a proto-relation describes the volume or area that another object would need to intersect to fulfill the associated relation with the anchor object. Proto-relations can encode information such as \enquote{somewhere next to the TV} or \enquote{the area behind the sofa.} This representation will arguably be useful for agents that use scene graphs as their intermediate knowledge state.

% we address these issues and as a consequence, provide a very good benchmark
\ourdata{} can be used to benchmark existing scene graph models to evaluate how suitable they are for downstream use. Previously, models had to be designed to cope with low-quality ground truth. Their potential with better ground truth could only be estimated. We train several scene graph generation models on our high-quality data and evaluate their performance.
%We argue that our benchmark is a good indicator for future scene graph performance.

Finally, we present a framework for reasoning on \ourdata{} scene graphs using small language models. This framework demonstrates the expressiveness of \ourdata{} when employed in downstream applications.

Our contributions can be summarized as follows:

\begin{enumerate}
   \item We present a new paradigm for relation representation, \textbf{parametric relations}, which eliminates imprecision of relation labels in the ground truth.
   \item We introduce \textbf{proto-relations}, a novel concept to encode hypothetical relations in the scene graph ground truth. Proto-relations can provide better reasoning and planning capabilities for downstream tasks.
   \item We design a pipeline to extract highly accurate and exhaustive scene graph ground truth from any 3D scene.
   \item Using our pipeline, we create \ourdata{}, a synthetic panoptic scene graph dataset with more than \numrel{} million relation annotations that is publicly available.
   \item We adapt a state-of-the-art scene graph generation model to predict parametric relations.
   \item For future reference, we perform a comparison and analysis of existing scene graph models on our dataset. % surprising results?
   \item Finally, we provide a convenient framework to query the extracted graphs for downstream applications using small language models that can run on-device.
\end{enumerate}

For the inference code and model weights, please see \url{https://anonymous.4open.science/r/paper-26E0}. The full code and dataset will be released upon paper acceptance.

\section{Related Work}

Our \ourdata{} dataset tackles fundamental issues that current scene graph generation methods face.
In this section, we analyze existing scene graph datasets, finding that they suffer from incomplete and inaccurate annotations. We introduce the scene graph models that we benchmark on \ourdata{} and list scene generators that can be paired with our approach.

\subsection{Scene Graph Datasets}
\label{sec:sg_datasets}

% this is a very important section of the paper
% we have to make sure to include all kinds of datasets and highlight how we are different
% this section will be an indirect proof of novelty for the paper

% 2D scene graph datasets

\begin{table}
   \begin{center}
   \resizebox{\linewidth}{!}{
   \begin{tabular}{lrcccr}
   \toprule
   Dataset & \#Relations & Seg & Depth & Normals & Coverage \\
   \midrule
   VG \cite{visual_genome} & 2.3M & \textcolor{Red}{$\times$} & \textcolor{Red}{$\times$} & \textcolor{Red}{$\times$} & 3.4\% \\
   AS-1B \cite{as1b} & 63k & \textcolor{Red}{$\times$} & \textcolor{Red}{$\times$} & \textcolor{Red}{$\times$} & 7.1\% \\
   PSG \cite{psg} & 275k & \textcolor{Green}{\checkmark} & \textcolor{Red}{$\times$} & \textcolor{Red}{$\times$} & 12.6\% \\
   Ours & \numrel{} & \textcolor{Green}{\checkmark} & \textcolor{Green}{\checkmark} & \textcolor{Green}{\checkmark} & 100.0\% \\
   \bottomrule
   \end{tabular}
   }
   \end{center}
   \caption{Comparison of scene graph datasets.
   \ourdata{} contains segmentation masks (Seg), depth maps (Depth), and surface normals (Normals).
   Coverage = how many of all the possible subject-object pairs are covered by at least one relation annotation.}
   \label{tbl:datasets}
\end{table}

We compare various scene graph datasets in \cref{tbl:datasets}.
Visual Genome \cite{visual_genome} was created by manually dividing each image into regions of interest with associated text captions. The scene graph was then extracted from the captions. Consequently, Visual Genome contains a high variety of relations but is rather sparsely annotated.
AS-1B \cite{as1b} focuses on the task of open-world panoptic visual recognition and understanding. The data generation process is very similar to Visual Genome except that large language models (LLMs) and vision-language models (VLMs) are employed instead of human annotators. Only a subset of AS-1B is human-verified.
\citeauthor{psg} \cite{psg} leverage the overlap of COCO \cite{coco} and Visual Genome to create PSG, a scene graph dataset with panoptic segmentation masks instead of bounding boxes. Compared to Visual Genome, the authors identify overlapping predicate class definitions and add a new condensed set of predicate classes. However, the annotation process is still heavily dependent on human annotators to remove ambiguous relations, which requires close scrutiny. Due to its manual annotation process, most relations still focus on salient relations, leading to a sparse scene graph.
In contrast, our \ourdata{} dataset provides complete scene graphs that are not confined to pre-established image regions. Instead, all relations are extracted using a well-defined set of rules that guarantee a consistent and reliable annotation.

\citeauthor{haystack} \cite{haystack} recognize the problem of incomplete scene graph annotations and propose to incorporate explicit negative ground truth in their Haystack dataset. This yields more reliable ground truth, but is still limited by a manual annotation process where guidelines sometimes provide imprecise predicate definitions and humans develop their own understandings in addition.
Similarly to Haystack, our dataset contains explicit negative ground truth but is much more exhaustively annotated.

3DSSG \cite{3dssg} is a semi-automatically generated scene graph dataset based on real-world scans of indoor environments \cite{3rscan}. The authors define a set of broad rules to extract relations from a scene. However, 3DSSG is based on noisy 3D data and human verification is still strictly required.
Our dataset is fully automatic and generates reliable ground truth at a much larger scale. Additionally, we introduce parametric relations, which provide much more fine-grained details like a distance instead of just saying that two objects are next to each other.

To assist the task of interactive indoor navigation, VLA-3D \cite{vla3d} provides 3D real-world scans from ScanNet \cite{scannet}, Matterport3D \cite{matterport3d}, Habitat-Matterport3D \cite{habitatmatterport3d}, 3RScan \cite{3rscan}, ARKitScenes \cite{aria_dataset}, and custom scenes generated with Unity \cite{unity}. The authors define a set of heuristics to automatically derive relation annotations from rotated object bounding boxes.
For \ourdata{}, we prioritize precise annotations. As such, we employ a voxel-based approach instead of bounding boxes and introduce the concept of parametric relations which are more flexible than a fixed set of heuristics.

\citeauthor{conceptgraphs} \cite{conceptgraphs} introduce ConceptGraphs, a pipeline for fusing multiple views of a 3D scene into a graph structure using LLMs and VLMs. The authors demonstrate multiple downstream applications for scene graphs.
For \ourdata{}, we follow a different route and provide the accurate and complete scene graphs as ground truth. Scene graph generation models can then learn to directly predict these graphs, leading to a more efficient graph generation pipeline. This is particularly beneficial for downstream applications that aim to perform processing directly on-device.

%With ConceptGraphs \cite{conceptgraphs}, there has been recent work to perform scene graph generation directly on 3D scenes. are an open-vocabulary graph representation for 3D scenes. To build, which uses object-centric 3D mapping system, combining geometric information from the traditional 3D mapping systems and semantic information from vision and language models. In ConceptGraphs, scene-objects are identified throuh a classs-agnostic segmentation model from the input RGB-D frames, later these objects are associated across multiple views based on geometric and semantic similarities. They used LLMs and VLMs to generate caption for the corresponding object and LLM for extracting inter-object relationships. Objects with 3D bounding box overlap  were only considered for constructing the relationship. However, in our dataset, we compute the relations following a motivation that, spatial relations potentially exists  even if there is no overlap between bounding boxes. Additionally, ConceptGraphs require human evaluators for performing scene graph evaluation because of the open-vocabulary setting, in contrast, our dataset doesn't demand human verification and uses LLMs for querying the generated scene graph.

\subsection{Scene Graph Models}

% now that we have criticised some existing scene graph datasets, we can say that our dataset is
% a good benchmark for existing scene graph models such as the ones described in this section

% don't spend too much on the individual methods (we are mostly focusing on data in this paper)
% Other 2D scene graph models that are out there. Acknowledge their innovative ideas but point out that in the end, accurate data matters most. That's what we're doing. We should analyse PSG methods but also standard SG methods. Maybe compare with LVM models too?

We evaluate several scene graph methods on our new dataset. Neural Motifs \cite{motifs} is a well-known scene graph model that is often used in comparisons. To boost performance, \citeauthor{motifs} propose to count relation occurrences in the dataset to derive a fixed bias. VCTree \cite{VCTree} is also a commonly used model that uses dynamic trees in combination with reinforcement learning to predict scene graphs.
% single-stage method
% OpenPSG here?
% DSFormer
Contrary to recent developments, DSFormer \cite{dsformer} was introduced as a true two-stage scene graph model. Since it directly leverages the segmentation information from the detection stage, it can focus on predicting accurate relations.
% SpeaQ
% \todo{Description of SpeaQ \cite{speaq} is missing.}

\subsection{3D Scene Generators}

% Not sure about this section, but we have to mention Infinigen somewhere.
% This section should be quite short. We can reference some other data generators like ProcThor to show
% that we know what we are doing.
Our pipeline can automatically extract precise annotations from any 3D data like ProcTHOR \cite{procthor} or Aria \cite{aria_dataset}.
To generate random scenes for \ourdata{}, we use Infinigen \cite{infinigen2023infinite,infinigen2024indoors} and combine it with our own automatic scene graph annotation pipeline. Infinigen is a rule-based data generator that is capable of creating complex and realistic looking scenes. Contrary to existing room layout generators \cite{aria_dataset,scenescript,procthor,scenenet} which place predefined assets in the scene, Infinigen procedurally generates every single object. This greatly increases the diversity of the generated data.

\section{\ourdata{} Dataset Construction}

In this section, we describe parametric relations and proto-relations that enhance the expressiveness of scene graphs. We also describe the creation of \ourdata{}.% and design a baseline model for parametric relation estimation.

\subsection{Parametric Relations}
\label{sec:param_rels}

% motivation
We introduce the concept of \textit{parametric} relations, which capture much more fine-grained information by representing a relation with a predicate class (\eg \textit{touching} or \textit{left of}) and a parameter (an angle or a distance) that more thoroughly describes the predicate.
For example, the parametric \textit{next to} relation %is a good example. 
%parametric relations do not constitute binary labels indicating whether two objects are next to each other. Instead, they
describes how close two objects actually are (\eg 50~cm). Traditional scene graphs, which only provide binary labels indicating whether two objects are next to each other, do not have %lose a lot of
this expressiveness. % by not providing any predicate parameters.

\textbf{Definition.}
For a given scene, we have a set of instances $I$, \eg chairs or windows. We define $P$ as the set of all predicate classes. Let $A \subset P$ be the set of all directional predicate classes (\eg \textit{in front of} or \textit{below}) and $D \subset P$ the set of all distance-based classes (\textit{next to} and \textit{touching}), with $A \cap D = \emptyset$. Predicate classes without parameters (\eg \textit{on}) are in $P \setminus (A \cup D)$. We then define a relation by a 6-tuple of subject instance $sbj \in I$, object instance $obj \in I$, predicate class $pred \in P$, relation parameter $\alpha \in \mathbb{R}$, camera perspective $cam \in \mathbb{R}^{3 \times 4}$ (the extrinsic matrix), and test direction $\vec{v} \in \mathbb{R}^3$.
$\vec{v}$ denotes the direction in scene space, along which the relation is determined.
% how to "read" a relation
A relation can be read as \enquote{$sbj$ - $pred$ - $obj$}, \eg \enquote{chair - in front of - window}. For \ourdata{}, we define the following predicate classes: \textit{in front of}, \textit{behind}, \textit{above}, \textit{below}, \textit{left}, \textit{right}, \textit{touching}, \textit{next to}, and \textit{on}.

% distance-based relations
\textbf{Distance-based relations.}
If the relation is a distance-based relation ($pred \in D$), the parameter $\alpha$ stores the shortest scene space distance between $sbj$ and $obj$ in meters.
%The relation direction $\vec{v}$ does not matter in this case and is ignored.
Note that distance-based relations are invariant to camera perspective. Thus, $cam$ can be ignored.
% how we extract the distance-based relations
To extract distances, we employ a voxel-based approach.
Prior work \cite{vla3d} uses 3D bounding boxes which only provide a rough estimation of the actual distance.
Our annotations are much more precise:
We first voxelise the 3D scene into a grid of 1~cm and then use a K-d tree to efficiently compute the shortest distance between the two voxelized objects. We define \textit{touching} as a special case of \textit{next to} with a distance of 0~m.
The granularity of the estimated distance is limited by the size of the voxel grid. Finer voxel grids are possible but come at an increased computational cost.

\begin{figure}
  \centering
   \includegraphics[width=\linewidth]{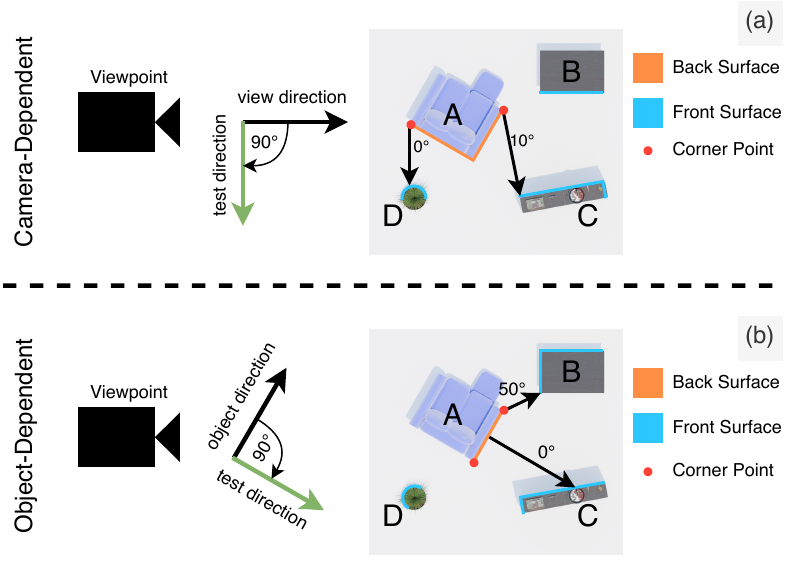}
   \caption{Schematic of the directional \textit{right of} relation. Shown are both the camera-dependent and the object-dependent variant. For a directional relation, we define the parameter $\alpha$ as the minimum angle between front and back surface compared to the test direction $\vec{v}$. The arrows show the connecting vectors from object to subject.}
   \label{fig:param_angle}
\end{figure}

% multiple views and the viewpoint index
\textbf{Directional Relations.}
For directional relations ($pred \in A$), camera perspective does matter.
Unlike most existing scene graph datasets, \ourdata{} provides multiple views of the same scene. Hence, two instances in the scene can have entirely different relations with each other, depending on the viewpoint.
%$\vec{v} \in \mathbb{R}^3$ denotes the direction in scene space, along which the relation is determined.
\cref{fig:param_angle}a illustrates the principle for the camera-dependent \textit{right of} relation. In this case, $\vec{v}$ (shown in green) is orthogonal to the forward vector of the camera as shown in the schematic. Other directional relations are defined analogously.

\begin{figure}
  \centering
   \includegraphics[width=\linewidth]{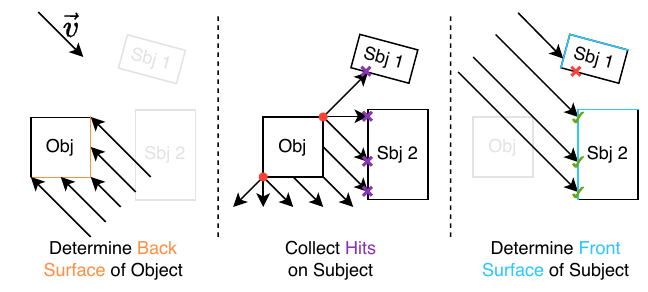}
   \caption{The ray sweep technique to determine back surfaces and front surfaces along the test direction $\vec{v}$. The hit on \textit{Sbj~1} is not on its front surface and the ray will be discarded.}
   \label{fig:raysweep}
\end{figure}

% how we extract directional relations
To extract a potential relation between $sbj$ and $obj$, we calculate the angle between front and back surfaces of $sbj$ and $obj$ along $\vec{v}$. This technique is depicted in \cref{fig:raysweep}.
% determine back surface of obj
To determine the back surface of $obj$, we perform a sweep in the opposite direction of $\vec{v}$ using ray casts in Open3D \cite{open3d}. The hit locations on $obj$ make up the back surface of $obj$.
% determine hits on surface of sbj
Next, we cast rays from the back surface along the $\vec{v}$ direction and determine if the rays hit $sbj$. On the corner points of the back surface (shown as red dots in \cref{fig:param_angle,fig:raysweep}), we additionally cast rays that deviate up to 90 degrees from $\vec{v}$. We make sure that the rays cover a resolution of 0.05~cm at a distance of 6~m.
% make sure that hits are on front surface of sbj
For each ray that hits $sbj$, we determine if the hit location is located on $sbj$'s front surface. To do so, we cast rays along the $\vec{v}$ direction onto the hit locations and check if the distance is below a small margin of error. If it is not, the hit is on $sbj$ but not on the front surface. We discard the associated rays as shown in \cref{fig:raysweep}.
% determining the final result
Finally, we collect all rays that hit the front surface of $sbj$ and keep the ray with the smallest angular deviation from $\vec{v}$. The angular deviation will be assigned to the parameter $\alpha$ of the associated relation.

% directional relations that are camera-independent
% When you explain the "prototypical" view of an object, it may be related to the tendency of humans to assign a "front" to objects.
% Then, your idea of a prototypical perspective "animates" the object and imagines looking into this direction.
% As discussed, this probably needs much more research (also maybe how to obtain this info automatically large-scale?). Here are some references (didn't read them, but please skim them + cite as you see fit when you talk about these definitions per relation: https://www.sciencedirect.com/science/article/pii/S0896627305000759#fig2
We also define directional relations that do not depend on any camera perspective $cam$. Humans tend to assign 3D orientation to certain objects \cite{front_definition}, \eg the \enquote{front} of cars is the direction into which they typically drive. Therefore, we identify a set of object categories \footnote{For the full list, see \cref{supp:mapping} in the supplementary} (\eg laptops or shelves) to which we can assign a \enquote{typical} orientation and define the test direction $\vec{v}$ based on the pose of $obj$ as depicted in \cref{fig:param_angle}b.
Intuitively, this captures the interpretation of the predicate class from the object's perspective.
See \cref{supp:indep_dirs} in the supplementary for some examples.
Using this interpretation vs.~the interpretation from the camera perspective is a common source for ambiguity in human scene relation interpretation.
With $\vec{v}$ determined, the extraction process is analogous to camera-dependent directional relations.
To differentiate between camera-dependent relations and camera-independent relations, we treat the associated predicates as different classes, \eg \textit{independent front} vs.~\textit{dependent front}.

% what is supposed to be predicted?
When predicting parametric relations using a scene graph generation model, only $sbj$, $obj$, $pred$, and $\alpha$ are relevant. $\vec{v}$ and $cam$ are only required during the extraction process to generate the dataset.

\subsection{Proto-Relations}
\label{sec:proto-rel}

\begin{figure}
    \centering
    \includegraphics[width=0.9\linewidth]{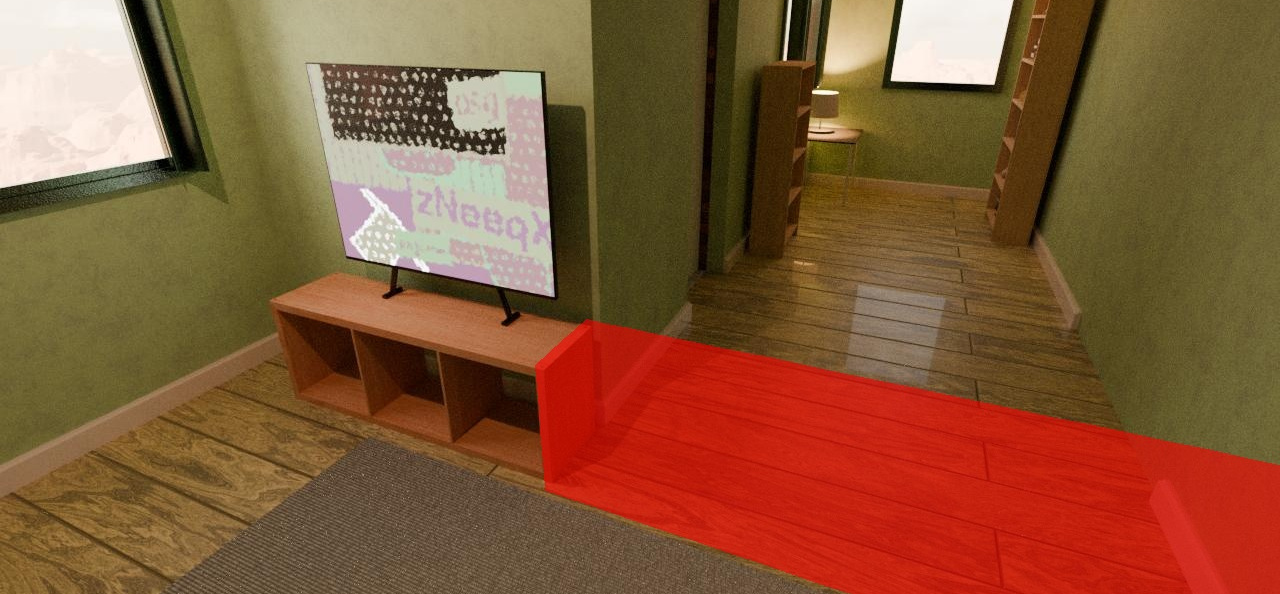}
    \caption{Example proto-relation (\textit{right of}) for the TV board. The area highlighted in red shows the associated volume for the proto-relation. Any new object placed inside the volume would fulfil a \textit{right of} relation with regard to the TV board with an angle deviation of 0 degrees.}
    \label{fig:proto_relation}
\end{figure}

We introduce proto-relations as a novel concept for scene graphs to capture hypothetical relationships. A proto-relation is defined as a triple of an anchor object in the scene, a predicate class label, and a volume or area in the scene. The proto-relation describes the fact that any new object intersecting with the volume would automatically fulfill the associated relation to the anchor object. Proto-relations enhance the expressiveness of scene graphs for downstream applications. They can be used to answer questions like \enquote{Where do I have to put this object such that it will be next to another one?} An example proto-relation can be seen in \cref{fig:proto_relation}.

\begin{figure}
    \centering
    \includegraphics[width=0.6\linewidth]{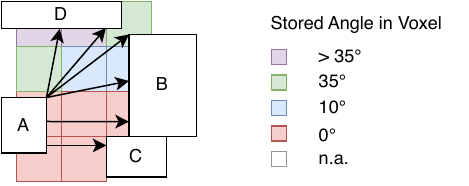}
    \caption{Schematic of the proto-relation extraction process. Similar to the extraction process for parametric relations, rays are cast in different directions. The smallest deviation from the test direction is stored in any voxel that is passed by a ray.}
    \label{fig:extract_proto}
\end{figure}

% motivation for proto-relations
%It is often argued that scene graphs can be used to efficiently convey information about an observed scene to a downstream application. However, these types of application are limited to relations between objects that exist in a scene. Traditional scene graphs cannot model hypothetical relationships to answer questions like "Where do I have to put this object such that it will be next to another one?".

% definition of proto-relations
%Proto-relations are triples of an anchor object in the scene, a predicate, and a volume in the scene. The proto-relation describes the fact that any new object that is placed inside the volume will automatically have the associated relation to the anchor object.
%For clarity, we refer to the volumes as proto-volumes in the following.

% how we extract proto-relations
To extract proto-relations, we follow the same extraction process as described in \cref{sec:param_rels}. However, instead of storing the object that is hit by the rays, we keep track of the space between ray origin and hit location. To this end, we divide the scene into 1~cm voxels and keep track of which voxels are crossed by a ray. As described in \cref{sec:param_rels}, we can determine a ray's deviation from the test direction $\vec{v}$ by an angle. Similar to parametric relations, we keep track of the angles of the rays crossing each voxel.
If one or more rays cross a voxel, we assign the smallest associated angle to the voxel as shown in \cref{fig:extract_proto}. Consequently, we end up with a volume of voxels where each voxel has an angle assigned to it. This enables us to filter the voxels by their angle deviation to again allow for different interpretations of relations similar to parametric relations.
% CSG
Proto-relations can be combined with constructive solid geometry (CSG) operations like intersection, difference, or union. See \cref{sec:application} for qualitative examples. Proto-relations and CSG operations provide an expressive foundation for downstream applications.
%
% predicting proto-relations will be future research
\ourdata{} contains proto-relations for every scene in OpenVDB \cite{openvdb} format for efficient access.
%We leave estimating proto-relations for arbitrary scenes to future work.

\subsection{Dataset Statistics}
\label{sec:stats}

For \ourdata{}, we create over 1.2k unique indoor scenes with entirely procedurally generated objects using Infinigen \cite{infinigen2024indoors}, accumulating more than 17 TB of raw ground truth data.
On average, we render 30 views per scene, totaling to 36k images. This provides sufficient variety while reducing the computation to generate scenes. Rendered examples are shown in \cref{fig:example_scene,fig:proto_relation}. For a subset of scenes, we render up to 200 views per scene to enable future research on multi-view evaluation of scene graphs. Each rendered image comes with segmentation masks, depth maps, and surface normals.
% object mapping
Since Infinigen does not assign class labels to the object instances in the scene, we create a custom mapping of object names to class labels (\cref{supp:mapping} in the supplementary).
Note that our pipeline does not depend on Infinigen and can be applied to any 3D dataset.

\begin{figure}
    \centering
    \includegraphics[width=\linewidth]{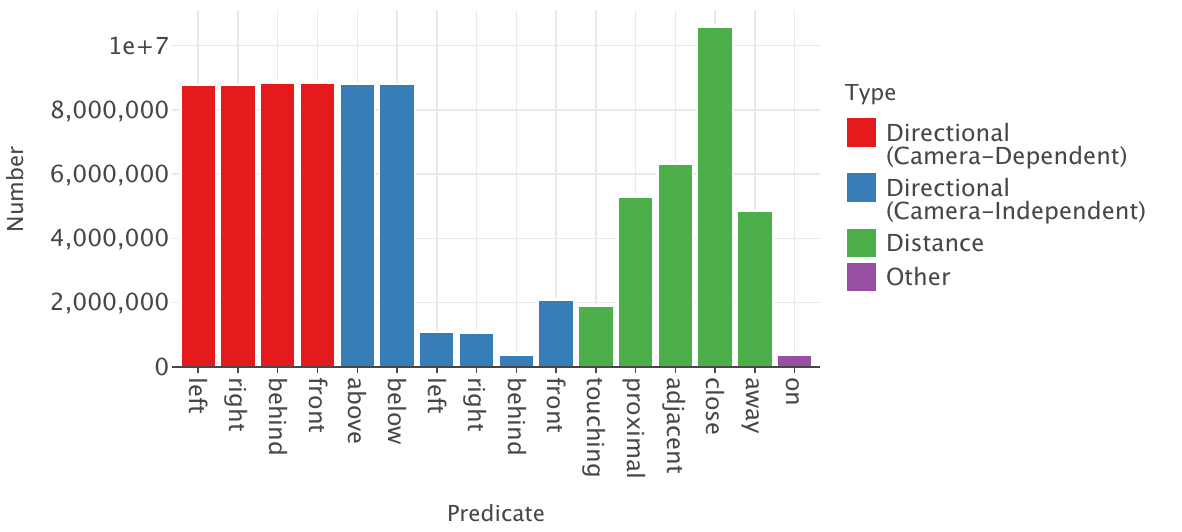}
    \caption{Predicate class distribution in the overall dataset. All distance based relations are stored internally as a single relation. This figure shows a discretized variant for clarity (proximal $\in (0, 0.3]$, adjacent $\in (0.3, 1]$, close $\in (1, 3]$, away $\in (3, \infty)$).}
    \label{fig:predicate-distr}
\end{figure}

% describe data stats. How much data do we have?
Using our exhaustive annotation process described in \cref{sec:param_rels,sec:proto-rel}, we extract more than \numrel{} relations with 2.4k relations per view and 72k relations per scene on average. \cref{fig:predicate-distr} shows the distribution of extracted predicate classes. For each relation and object instance, we provide the corresponding proto-relation as described in \cref{sec:proto-rel}.

Compared to existing datasets, we provide an exhaustive annotation of the full scene, whereas other datasets only annotate 3.4\% to 12.6\% of all possible subject-object pairs as shown in \cref{tbl:datasets}.

We split the data into 60\% training data, 10\% validation data, and 30\% test data. We make sure that all images that belong to the same scene are in the same data split.

\section{Scene Graph Generation with Parametric Relations}
\label{sec:baseline}

In this section, we illustrate how to extend scene graph models to capture parametric relations.
%We extend DSFormer \cite{dsformer} to capture
%and show how to extend a scene graph model for parametric relations.
% why DSFormer is a good choice (2-stage!)
As a two-stage method, DSFormer \cite{dsformer} is a good candidate to analyze scene graph capabilities on a novel dataset, because it separates object detection from relation detection. Consequently, the reported relation results do not depend on object detection performance.
Contrary to most single-stage methods which have a limit for the total number of relations per image, two-stage methods can usually parse all relations.
DSFormer is designed to return multi-label predictions for relations between two objects, which is necessary for modeling \ourdata{}, which contains many relations with different predicate classes for the same subject-object pairs.
%It is sufficient to only modify the final linear layer by doubling the number of outputs to accommodate parameter predictions. For every parameter output that is associated with an angle, we apply a sigmoid function after the output to ensure values in $[0, 1]$. These values are then interpreted as the cosine of the angle.
During training, we initialize the backbone with pretrained weights from DINOv2 \cite{dinov2} and freeze it. We then train the remaining layers.

% modifications to DSFormer and training
%To predict parametric relations, a scene graph model must return two outputs: (1) whether the relation exists at all (we will call these outputs flags) and if yes, (2) the parameter of the relation. Therefore, we make sure that the final layer returns two outputs per predicate class. It is sufficient to only modify DSFormer's final layer by doubling the number of outputs.
Assume a set of parametric relations defined by $n$ predicate classes.
We define $A \subset [1, ..., n]$ as the set of angle-based predicate classes and $D \subset [1,...,n]$ as the set of distance-based predicate classes.
To predict whether a parametric relation exists between a subject-object pair, a scene graph model must return two outputs: (1) a binary output flag $\hat{f}$ indicating whether the relation exists at all and if it does, (2) the parameter of the relation $\hat{p} \in \mathbb{R}$.
%To ensure that the final layer returns two outputs per predicate class,
%It is sufficient to
We modify DSFormer's final multi-label prediction layer % by doubling the number of outputs.
such that it returns output flags $\hat{f} \in [0,1]^n$, and output parameters $\hat{p} \in \mathbb{R}^n$ for each input subject-object pair, predicting all potential relations simultaneously.

% mathematical formulation
%For a single subject-object pair, we define the number of predicate classes $n$, binary ground truth flags $f \in \{0,1\}^n$, ground truth parameters $p \in \mathbb{R}^n$, binary output flags $\hat{f} \in [0,1]^n$, and output parameters $\hat{p} \in \mathbb{R}^n$. Additionally, we define $A \subset [1, ..., n]$ as the set of angle-based predicate classes and $D \subset [1,...,n]$ as the set of distance-based predicate classes.

During training, we make use of ground truth flags $f \in \{0,1\}^n$ and ground truth parameters $p \in \mathbb{R}^n$.
To train the flags $\hat{f}$, we employ a binary cross entropy loss with $f$ as described in \cite{dsformer}.
To train the parameter outputs $\hat{p}$, we introduce parameter-specific loss terms for angles ($\mathcal{L}_A$) and distances ($\mathcal{L}_D$):

\begin{align}
    \mathcal{L}_A(f, p, \hat{p}) &= \frac{1}{\sum_{i \in A} f_i} \sum_{i \in A} f_i \cdot \left(cos(p_i) - \hat{p}_i\right)^2 \\
    \mathcal{L}_D(f, p, \hat{p}) &= \frac{1}{\sum_{i \in D} f_i} \sum_{i \in D} f_i \cdot \frac{smooth_{L_1}(p_i - \hat{p}_i)}{p_i + 1}
\end{align}

$smooth_{L_1}$ is the smooth L1 loss \cite{fast_rcnn}. The distance loss is divided by $p_i + 1$ to focus more on closer distances.
%All angles are expected to be in radians.
For all angle parameters, we apply the sigmoid function to the respective output before calculating $\mathcal{L}_D$ such that $\hat{p}_a \in [0,1]\; \forall a \in A$. This allows the model to predict angles as cosine values for angles in the range $\left[0, \frac{\pi}{2}\right]$.
% why no cosine on model output
%For angle parameters, we don't apply 

\section{Experimental Results}

We evaluate scene graph generation methods on \ourdata{}, introducing new metrics and conducting experiments focusing on parametric relations and multi-view aggregation.

\subsection{Evaluation Metrics}
\label{sec:metrics}

To evaluate on the non-parametric variant of \ourdata{}, we use No Graph Constraint Mean Recall@k (\textit{ng-mR@k}) and Mean Average Precision (\textit{mAP}). If parametric relations are used, we employ \textit{mAP} and Mean Absolute Error.

% mR@k doesn't work with our dataset, explain why
\textbf{Non-parametric metrics.}
Traditionally, because existing datasets are not exhaustively annotated with relations, scene graph models are evaluated using metrics such as Recall@k \cite{Rk_definition} and Mean Recall@k (\textit{mR@k}) \cite{kern,VCTree} with $k \in \{20,50,100\}$. However, a perfect model could only achieve a \textit{mR@50} of 0.077 on our dataset. The reasons are twofold: (1) \ourdata{} is a true multi-label dataset but \textit{mR@k} only allows one predicate per subject-object pair. (2) Our dataset is much more densely annotated (see \cref{sec:stats} for more information). Choosing only the top 50 predicted relations does not cover most of the ground truth relations. Therefore, we increase $k$ to 1000 and adopt \textit{ng-mR@k} \cite{pixel2graph,motifs}, which is a variation of \textit{mR@k} that allows multiple predicates per relation.

% argue why we can now use average precision
The \textit{R@k} metrics were originally introduced because traditional scene graph datasets only contain positive relation annotations, \ie when a relation exists \cite{first_scenegraph}. There have been attempts to integrate negative relation annotations in scene graph datasets \cite{haystack}, but they were limited to a small dataset size.
With our dataset, we provide exhaustively annotated ground truth relations and are no longer limited to \textit{R@k} metrics. Moreover, we can %finally
adopt metrics %that are commonly used in other domains, such as object detection. We suggest to use
such as average precision (\textit{AP}) per predicate class, which in contrast to \textit{R@k} captures all predictions for a scene and produces more reliable values even for imbalanced data distributions. We additionally compute the \textit{mAP} as the average over all predicate scores.

% parametric relation metrics
\textbf{Parametric metrics.}
We break down evaluation into two distinct tasks: (1) predicting relation existence as a binary label and (2) predicting the value of the parameter if the relation exists. For the existence-task, we again employ \textit{AP} as we did for the non-parametric variant. For parameter value prediction, we report the mean absolute error per predicate class for all instances carrying the predicate class label in the ground truth regardless of whether the model predicts the relation to hold.
%If a relation exists but the model says otherwise, we still include the parameter prediction in the evaluation.
%If a relation does not exist in the ground truth, the parameter value is ignored.
%To compare two models, both metrics should always be considered.

\subsection{Performance with Parameter Thresholds}

\begin{table}
    \begin{center}
        \begin{tabular}{lrrrr}
            \toprule
            & \multicolumn{2}{c}{PredCls $\uparrow$} & \multicolumn{2}{c}{SGDet $\uparrow$} \\
            \cmidrule(lr){2-3}\cmidrule(lr){4-5}
            Method & mAP & ng-mR & mAP & ng-mR \\
            \midrule
            %MotifNet \cite{motifs} & 0.645 & 0.347 \\
            MotifNet \cite{motifs} & 64.1 & 34.4 & 27.0 & 24.4 \\
            % VCTree \cite{VCTree} & 0.644 & 0.348 \\
            VCTree \cite{VCTree} & 63.9 & 34.5 & 27.5 & 24.9 \\
            % OpenPSG \cite{openpsg_model} & ?? & ?? \\
            % SpeaQ \cite{speaq} & - & - & ?? & 22.7 \\
            DSFormer \cite{dsformer} & 67.0 & 45.4 & 30.7 & 30.5 \\
            %DSFormer + Param (ours) & ?? & ?? \\
            \bottomrule
        \end{tabular}
    \end{center}
    \caption{Model comparison on \ourdata{} with parameter thresholds. mAP is the mean average precision. For ng-mR, we use $k = 1000$. We evaluate on predicate classification (\textit{PredCls}) and scene graph detection (\textit{SGDet}).
    % SpeaQ does not support evaluation on \textit{PredCls}.
    Each method was retrained on our dataset.
    }
    \label{tbl:threshperf}
\end{table}

\begin{table*}
    \begin{center}
    \resizebox{\linewidth}{!}{
\begin{tabular}{llcccccccccc}
\toprule
\multicolumn{2}{c}{} & \multicolumn{6}{c}{Camera Independent} & \multicolumn{2}{c}{Camera Dependent} & \multicolumn{2}{c}{Average} \\
\cmidrule(lr){3-8}\cmidrule(lr){9-10}\cmidrule(lr){11-12}
Method & Metric & front & above & right & distance & on & touching & front & right & Angle & All \\
\midrule
RGB + 1 Block & AP $\uparrow$ & 74.41 & 82.17 & 44.43 & - & 65.44 & 73.92 & 77.94 & 64.65 & 68.72 & 68.99 \\
RGB + 1 Block & Parameter Error $\downarrow$ & 22.77\textdegree & 12.66\textdegree & 23.30\textdegree & 1.15 m & - & - & 15.59\textdegree & 17.85\textdegree & 18.44\textdegree & - \\
RGB + 4 Blocks & AP $\uparrow$ & 87.74 & 89.32 & 54.39 & - & 81.00 & 82.93 & 87.81 & 70.18 & 77.89 & 79.05 \\
RGB + 4 Blocks & Parameter Error $\downarrow$ & 17.93\textdegree & 10.82\textdegree & 19.71\textdegree & 1.15 m & - & - & 13.17\textdegree & 15.56\textdegree & 15.44\textdegree & - \\
Depth + 4 Blocks & AP $\uparrow$ & 86.95 & 89.74 & 53.59 & - & 81.14 & 83.14 & 88.94 & 70.16 & 77.87 & 79.09 \\
Depth + 4 Blocks & Parameter Error $\downarrow$ & 17.37\textdegree & 10.73\textdegree & 19.32\textdegree & 1.15 m & - & - & 12.54\textdegree & 14.88\textdegree & 14.97\textdegree & - \\
\bottomrule
\end{tabular}
}
    \end{center}
    \caption{Results for the proposed parametric parametric scene graph model. The table omits the \textit{behind}/\textit{below}/\textit{left} predicates since they perform very similar to their \textit{front}/\textit{above}/\textit{right} counterparts.}
    \label{tbl:paramperf}
\end{table*}

\ourdata{} provides an extensive benchmark environment for existing scene graph models. Using our dataset, authors can now reliably evaluate the capabilities of their models on exhaustive ground truth annotations.
To use \ourdata{} with prior work, we introduce a variant of our dataset that applies thresholds to the relation parameters. Every relation with an angle parameter $\le 10$\textdegree\ counts as positive and every relation with an angle parameter $> 20$\textdegree\ counts as negative. Everything between will be ignored during evaluation. For the \textit{next to} relation, distances $\le$ 1~m count as positive, distances $>$ 1.2~m count as negative. We train each of the compared methods on the same training split of our dataset and evaluate using the Predicate Classification (\textit{PredCls}) and Scene Graph Detection (\textit{SGDet}) protocols. In \textit{PredCls}, the model receives the ground truth segmentation masks in the image as input and predicts the relations between them. In \textit{SGDet}, the model only receives the image and has to identify the instances on its own.

% interpretation
The results can be found in \cref{tbl:threshperf}. On our benchmark, DSFormer is the clear winner, with a mAP of 0.670 in \textit{PredCls} and 0.307 in \textit{SGDet}. Interestingly, MotifNet \cite{motifs} outperforms VCTree \cite{VCTree} with \textit{PredCls} on \ourdata{} even though it performs consistently worse on the Visual Genome \cite{visual_genome} and PSG \cite{psg} benchmarks. As shown in \cref{tbl:threshperf}, ng-mR@1000 scores are not necessarily tied to mAP scores.

\subsection{Estimation of Parametric Relations}

We report results for our proposed method for scene graph generation with parametric relations in \cref{tbl:paramperf}.
All the compared models receive segmentation masks as input, but not the object categories. This evaluation protocol is very similar to \textit{SGDet} except that we also evaluate parameters.
Compared to the original DSFormer, we replace the ResNet-50 backbone with a ViT-S \cite{vit}, pretrained using DINOv2 \cite{dinov2} for increased performance. We train the models on the training set of \ourdata{} and show the results on the test set. To test the required model complexity for the task, we train with a transformer module of 1 and 4 blocks. In our experiments, the smaller model trains almost 5 times faster but achieves worse scores, especially on the relation existence check (0.69 versus 0.79 mAP).
In addition to a standard training on RGB images, we also train with depth-only images from \ourdata{}.

Across all evaluated methods, the \textit{above} relation can be predicted best. This is mainly because objects that are on top of each other typically don't occlude each other.
We also observe that camera-independent \textit{front} and \textit{right} relations are more difficult to predict than their camera-dependent counterparts. Predicting camera-independent relations requires knowledge about the pose of the object, making it a more complex task.
In general, \textit{front} appears to be a simpler task than \textit{right}. Many objects that are in front of each other are close (\eg shelves in front of walls and items in shelves). Hence, the model needs to focus only on a smaller region to grasp this relation.
\begin{figure}
    \centering
      \begin{subfigure}[c]{0.45\linewidth}
     %     \centering
          \includegraphics[width=\linewidth]{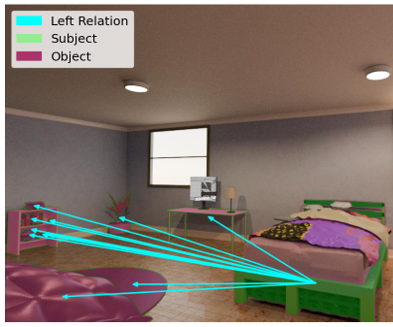}
      \end{subfigure}
      \hfill
     \begin{subfigure}[c]{0.45\linewidth}
     %     \centering
         \includegraphics[width=\linewidth]{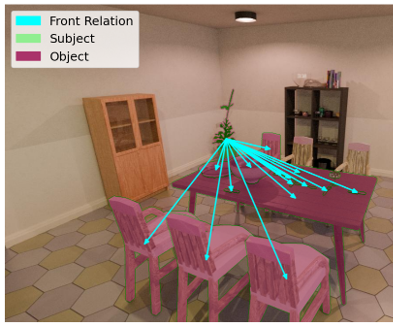}
     \end{subfigure}
     % \begin{subfigure}[c]{0.48\linewidth}
     %     \centering
         % \includegraphics[width=\linewidth]{img/front_30_d6c52a6_Image_3_0_0048_0.png}
     % \end{subfigure}
     % \hfill
     % \begin{subfigure}[c]{0.48\linewidth}
     %     \centering
         % \includegraphics[width=\linewidth]{img/e5d4551_Image_3_0_0048_0.jpg}
     % \end{subfigure}
    \caption{Qualitative prediction results on \ourdata{}.
    %Additional examples can be found in \cref{supp:qual_results}.
    }
    \label{fig:qual_results}
\end{figure}
The RGB model and depth model exhibit similar performance, though the depth model consistently achieves lower parameter errors.
This observation aligns with expectations since \ourdata{} includes spatial relations, and depth maps provide spatial information to the model.

% The RGB model with 4 blocks outperforms the depth model on the relation existence task (measured in AP) by a small margin. We attribute this to the fact that certain combinations of subject-object categories are more likely to form a relation. For example, chairs are more likely to be standing on the floor than windows. The depth model does not recognize the categories as easily.

\subsection{Multi-View Evaluation}

% what is multi-view evaluation
Since \ourdata{} contains multiple different views per scene, we can perform multi-view evaluation. We use a scene graph model and predict a separate scene graph for each view.
% how is it exactly calculated
% 1. calculate scene graph per view (only keep camera-independent)
% 2. pick random subset of views
% 3. aggregate predictions by taking the median over the outputs
% 4. evaluate
Since we want to evaluate the same relations from various views, we limit ourselves to camera-\emph{independent} relations.
We aggregate the predictions from the different views by taking the median over the model outputs.
Note that our multi-view evaluation is similar to that presented by \citeauthor{multiview} \cite{multiview} but instead of trying to associate object recognitions from different viewpoints, we evaluate how the detection of various relations can benefit from additional viewpoints.

% results + interpretation
\cref{tbl:multiview} shows the performance when aggregating multiple views. Using the outlined approach, performance saturates at around 15 views per scene with an increase in AP of up to +6.2.
%The performance increase is dependent on the type of relation.
The least impact can be observed for \textit{above} and \textit{below} since they have already high scores for a single view.
The \textit{front} relation benefits most from multiple views which might be because \textit{front} is easy to detect if a view aligns with the object orientation.
%For directional relations (\textit{behind}, \textit{front}, \textit{right}, \textit{left}, \textit{above}, \textit{below}), AP increases by $0.03$ - $0.06$.
%Distance-based relations (\textit{touching}, \textit{on}), appear to benefit more from multiple views with an AP increase of $0.05$ and $0.1$.

\begin{table}
   \begin{center}
   \resizebox{\linewidth}{!}{
\begin{tabular}{lrrrrrrrr}
\toprule
\# Views & behind & front & right & left & above & below & touching & on \\
\midrule
1 & 33.8 & 65.9 & 30.2 & 28.3 & 86.2 & 85.6 & 78.1 & 76.8 \\
\hline
\multicolumn{9}{c}{Improvement over single view} \\
\hline
2 & + 0.9 & + 1.6 & + 0.9 & + 1.8 & + 1.2 & + 1.0 & + 1.6 & + 1.6 \\
5 & + 4.8 & + 5.7 & + 4.0 & + 4.6 & + 2.8 & + 2.9 & + 3.4 & + 4.7 \\
10 & + 5.3 & + 6.1 & + 4.5 & + 4.9 & + 3.1 & + 3.1 & + 3.7 & + 5.1 \\
15 & + 5.6 & + 6.2 & + 4.6 & + 5.1 & + 3.2 & + 3.2 & + 3.8 & + 5.3 \\
All & + 5.6 & + 6.2 & + 4.6 & + 5.1 & + 3.2 & + 3.2 & + 3.8 & + 5.3 \\
\bottomrule
\end{tabular}
   }
   \end{center}
   \caption{AP scores in \% per camera-independent predicate when aggregating across different views of the same scene. The values show the increase compared to a single view. Aggregation is restricted to views where both the subject and object relevant to the respective relation are visible. Predictions are aggregated using the median of the model outputs.}
   \label{tbl:multiview}
\end{table}

% why is it relevant?
Especially for downstream applications that involve agents which can move around in a scene, multi-view predictions are arguably beneficial.
% limitations and future work
In this work, we have focused on camera-independent relations. Future work should further explore how to reconcile full scene graphs from different perspectives.

% \subsection{Transfer to Real Data}
% \label{sec:result_depth}

% To demonstrate the impact of \ourdata{} on real-world applications, we train a scene graph model on our dataset and evaluate it on PSG \cite{psg}. For this demonstration we discard RGB data as input because the domain gap is too large. However, depth and surface normals can be easily transferred. We train DSFormer on depth-only frames. When evaluating on PSG, we generate pseudo-depth using Depth Anything v2 \cite{depth_anything_v2} and use the pseudo-depth as input for our model. Thus, the full model pipeline is capable of processing RGB frames.

\subsection{Reasoning on \ourdata{}}
\label{sec:application}

\begin{figure}
    \centering
    \includegraphics[width=.9\linewidth]{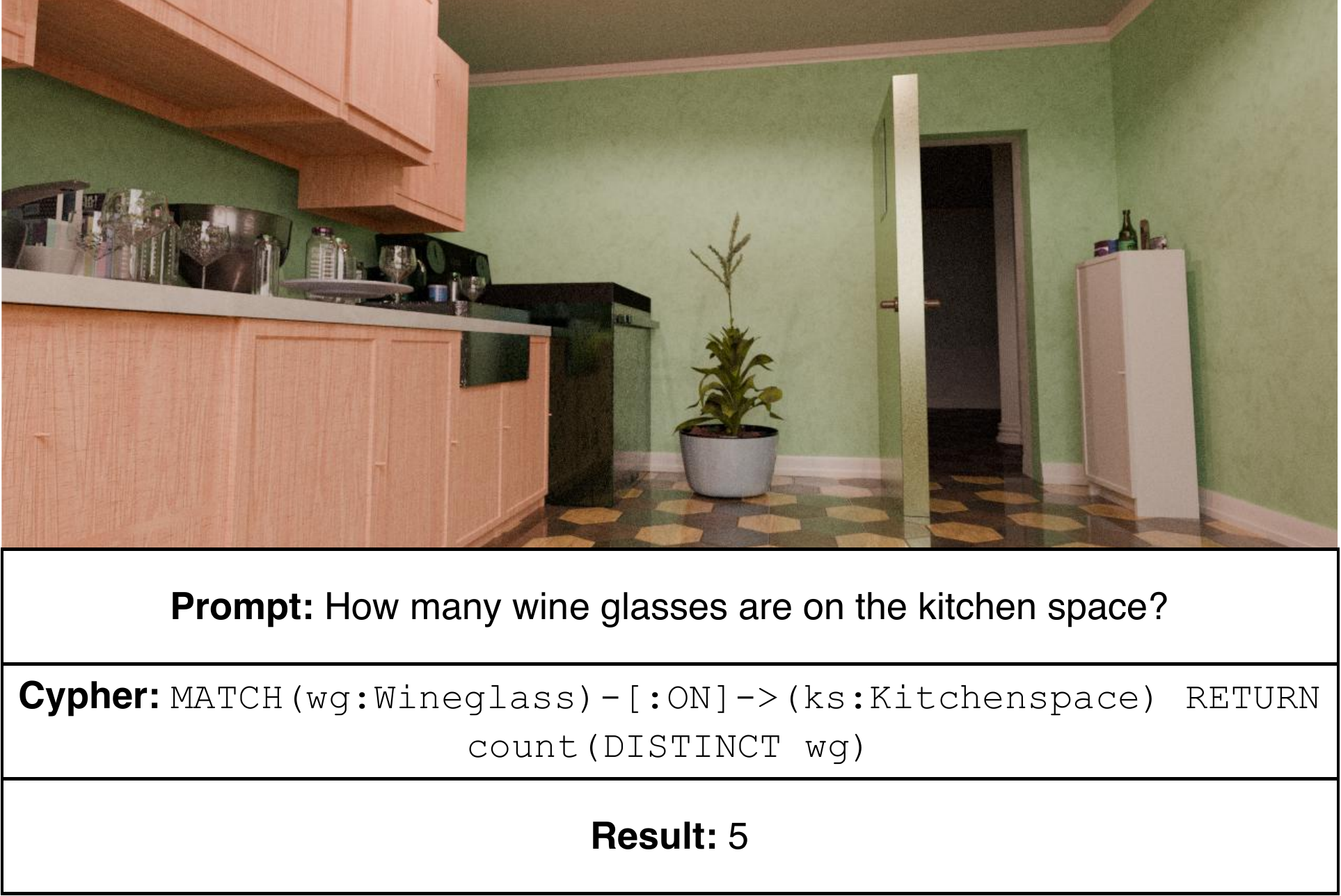}
    \caption{Example interaction with Qwen2.5 \cite{qwen2.5} on our dataset. The LLM returns a Cypher query which is executed to return the final results.}
    \label{fig:llm_reasoning}
\end{figure}

% scene graph queries:
% - user provides input prompt
% - convert scene graph into graph database
% - input prompt is converted to a graph query
% - graph query is applied to the graph database
To highlight possible applications of \ourdata{}, we provide the code for an easy to use scene graph query framework. With this framework, users can conveniently query information that is encoded in the graph. First, we convert the predicted scene graph into a Neo4j \cite{neo4j} graph database. Next, we employ a small language model (\eg Phi-4 \cite{phi4} or Qwen2.5:14b \cite{qwen2.5}) to generate Cypher \cite{cypher} graph queries based on the user's prompt. Finally, the graph query is applied to the graph database and the result is returned. \cref{fig:llm_reasoning} shows a qualitative example of this approach.

\begin{figure}
    \centering
    \includegraphics[width=\linewidth]{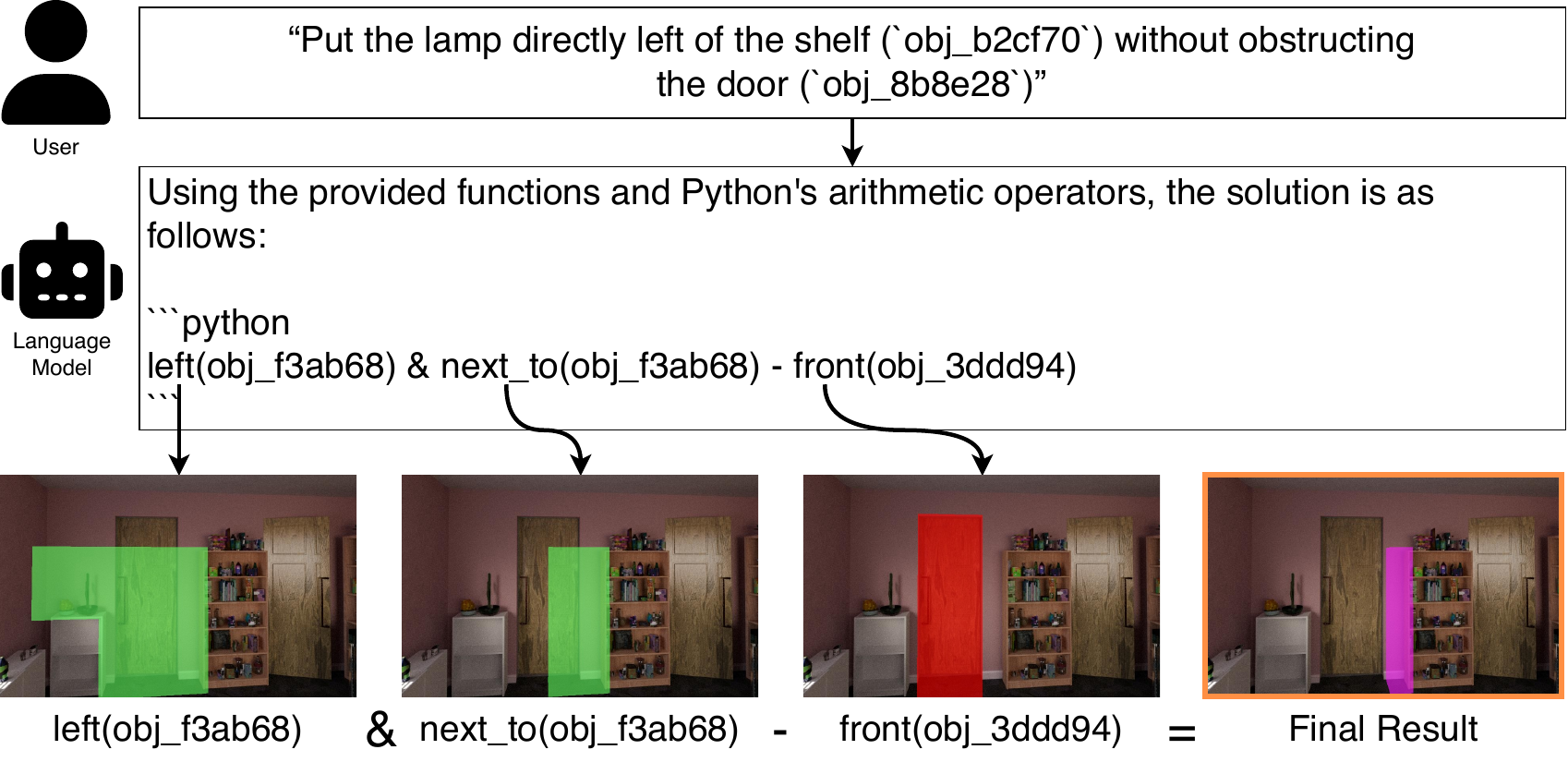}
    \caption{Example interaction with Phi4 \cite{phi4} for proto-relation queries. The language model returns a subset of Python that is interpreted as set of constructive solid geometry operations.}
    \label{fig:llm_proto}
\end{figure}

% proto volumes:
% - user provides input prompt
% - LLM converts input prompt to a set of CSG operations
% - CSG operations are applied to proto volumes
% - resulting volume is returned
Additionally, we provide a query framework for proto-relations to showcase the capabilities of the concept. Given a user prompt, we use a small language model to come up with a set of fitting boolean operations. The model is instructed to generate Python code with a restricted set of available functions, each associated with a proto-relation category (\eg \textit{in front of}, \textit{touching}).
See \cref{fig:llm_proto} for a qualitative example.
To query proto-relations, small language models are sufficient which enables on-device processing.

\section{Conclusion}

We have introduced \ourdata{}, a new synthetic scene graph dataset with exhaustive relation annotations per scene. Contrary to traditional scene graphs, which focus only on salient relations, our proposed annotation pipeline generates complete and precise scene graph annotations. \ourdata{} contains more than \numrel{} relations.

We design two new fundamental relation representations, parametric relations and proto-relations.
% parametric relations
\textbf{Parametric relations} represent relations not only with a predicate class but with an additional parameter which can be either an angle or a distance. These parametric relations tackle the problem of imprecise labels in scene graphs by quantifying them exactly instead of relying on a black box annotation process that simply assigns a binary label.
% baseline model
%We provide a baseline model to estimate parametric relations and evaluate several scene graph models on \ourdata{}.
% proto-relations
\textbf{Proto-relations} further extend traditional scene graphs by providing means to represent hypothetical relations which was not possible before.
% Proto-relations describe a volume where any intersecting object would satisfy the associated relation criterion.
% This can be used by downstream applications to assist in reasoning and planning tasks.
% our application framework
We provide a query framework and show potential applications using our proposed ground truth.

We believe that scene graphs can offer a unique representation of the world for more complex systems. Our two newly introduced relation types provide essential foundations for an effective use of scene graphs in downstream applications. We believe that accurate and expressive data as contained in \ourdata{} is a requirement for future research in this direction.
% Future research should explore new ideas how to effectively estimate parametric relations and proto-relations and how to employ them in downstream applications.

{
    \small
    \bibliographystyle{ieeenat_fullname}
    \bibliography{main}
}

\clearpage
\setcounter{page}{1}
\maketitlesupplementary

\def\ourdata{\textsc{CoPa-SG}}

\section{Directional Relations}
\label{supp:indep_dirs}

\begin{figure}[h]
    \centering
    \includegraphics[width=\linewidth]{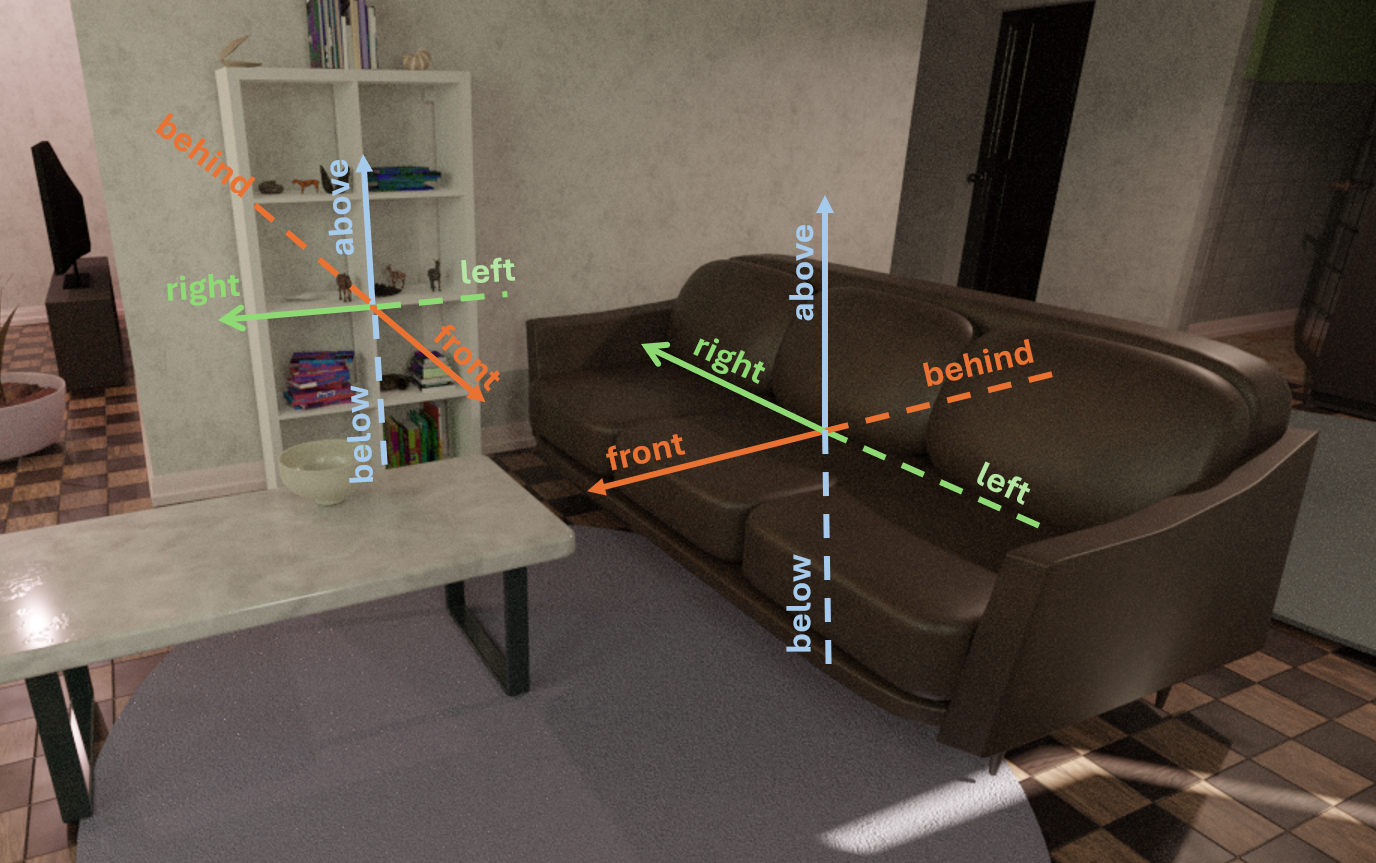}
    \includegraphics[width=\linewidth]{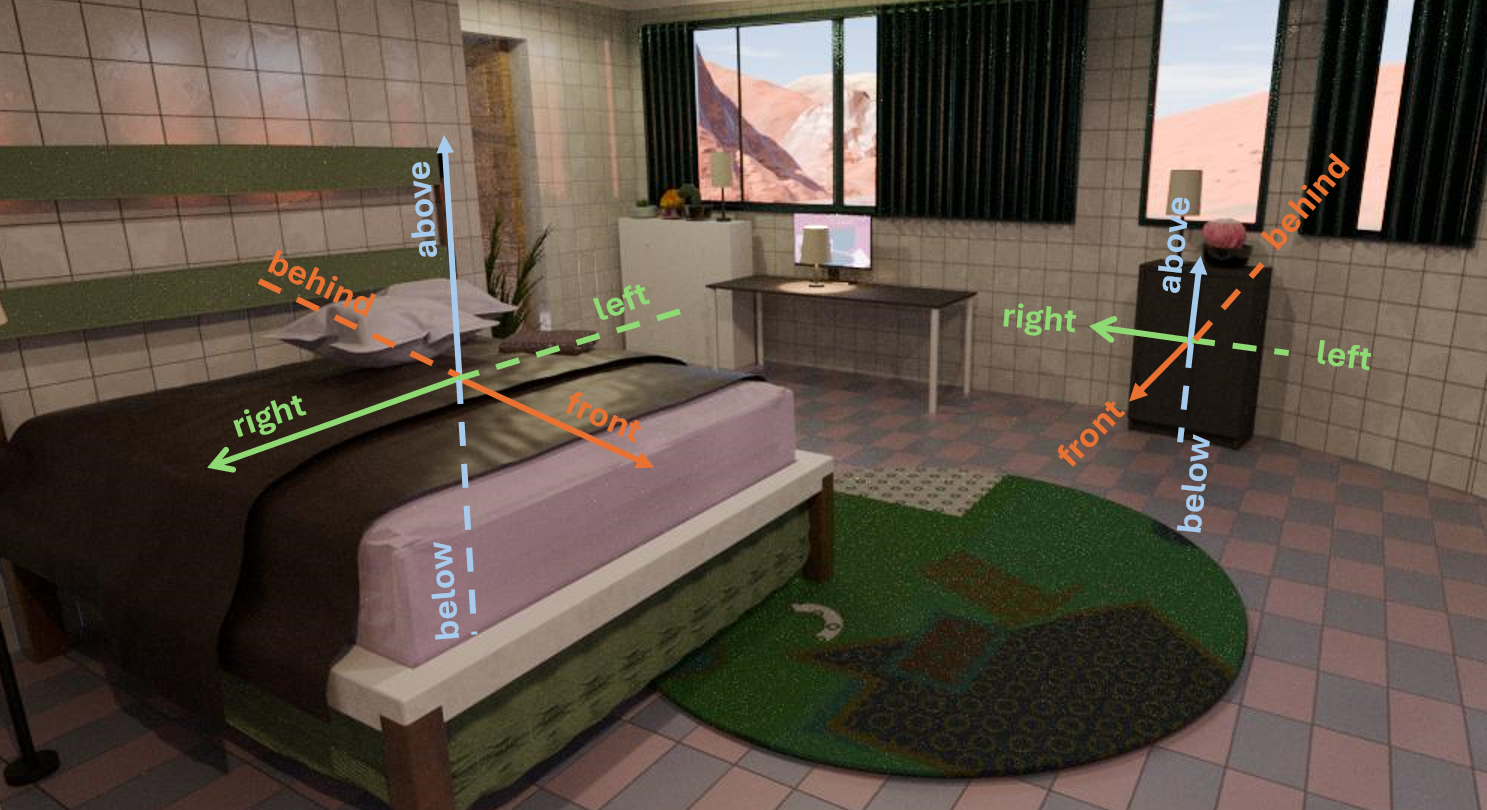}
    \caption{Test directions for camera-independent relations that depend on the object pose.}
    \label{fig:indep_dirs}
\end{figure}

Camera-independent relations are defined based on the pose of the associated object. \cref{fig:indep_dirs} shows how the various directions are defined with respect to the object's pose.

% yaaaaaaaaay!
% can we make the arrows/text a bit bigger?
% and the lines a bit thicker maybe i will crop it a bit more..arrows can be made more thick !!

\vfill

\pagebreak

\section{Infinigen Object Mapping}
\label{supp:mapping}

Since Infinigen \cite{infinigen2024indoors} does not include a set of class labels for the various objects, we create a mapping from object instance name to class label. You can see the mapping in \cref{tbl:infinigen_names1,tbl:infinigen_names2}. The \enquote{Directional} column specifies whether the respective object can serve as an object for a camera-independent directional relation.

\begin{table}[h]
    \centering
\begin{tabular}{lll}
\toprule
Infinigen Name & Class Label & Directional \\
\midrule
Balloon & balloon & No \\
BarChair & chair & Yes \\
Bathtub & bathtub & No \\
Bed & bed & Yes \\
BeverageFridge & fridge & Yes \\
Blanket & blanket & No \\
BlenderRock & rock & No \\
BookColumn & book & No \\
BookStack & book & No \\
Bottle & bottle & No \\
Bowl & bowl & No \\
BoxComforter & blanket & No \\
CeilingLight & light & No \\
CellShelf & shelf & Yes \\
Chair & chair & Yes \\
Chopsticks & cutlery & No \\
CoffeeTable & table & No \\
Comforter & blanket & No \\
Cup & cup & No \\
DeskLamp & light & No \\
Dishwasher & dishwasher & Yes \\
FloorLamp & light & No \\
Fork & cutlery & No \\
Fruit & fruit & No \\
GlassPanelDoor & door & No \\
Hardware & hardware & No \\
KitchenCabinet & shelf & Yes \\
Knife & cutlery & No \\
LargePlantContainer & plant & No \\
LargeShelf & shelf & Yes \\
LiteDoor & door & No \\
LouverDoor & door & No \\
\bottomrule
\end{tabular}
\caption{Mapping of Infinigen name to class label.}
\label{tbl:infinigen_names1}
\end{table}

\pagebreak

\begin{table}[h]
    \centering
    \begin{tabular}{lll}
\toprule
Infinigen Name & Class Label & Directional \\
\midrule
Mattress & mattress & No \\
Microwave & microwave & Yes \\
Mirror & mirror & Yes \\
Monitor & screen & Yes \\
NatureShelfTrinkets & trinket & No \\
OfficeChair & chair & Yes \\
Oven & oven & Yes \\
Pan & pan & No \\
PanelDoor & door & No \\
Pillar & pillar & No \\
Pillow & pillow & No \\
PlantContainer & plant & No \\
Plate & plate & No \\
Pot & pot & No \\
Rug & rug & No \\
SideTable & table & No \\
SimpleBookcase & shelf & Yes \\
SimpleDesk & table & No \\
SingleCabinet & shelf & Yes \\
Sink & sink & Yes \\
Sofa & sofa & Yes \\
Spoon & cutlery & No \\
StandingSink & sink & Yes \\
TV & screen & Yes \\
TVStand & shelf & Yes \\
TableDining & table & No \\
Toilet & toilet & Yes \\
Towel & towel & No \\
Vase & vase & No \\
WallArt & art & Yes \\
Window & window & Yes \\
Wineglass & wineglass & No \\
\bottomrule
\end{tabular}
\caption{Mapping of Infinigen name to class label.}
\label{tbl:infinigen_names2}
\end{table}

\pagebreak

\section{Predicate Distribution in \ourdata{}}

\begin{figure}[h]
    \centering
    \begin{subfigure}[c]{\linewidth}
        \includegraphics[width=\linewidth]{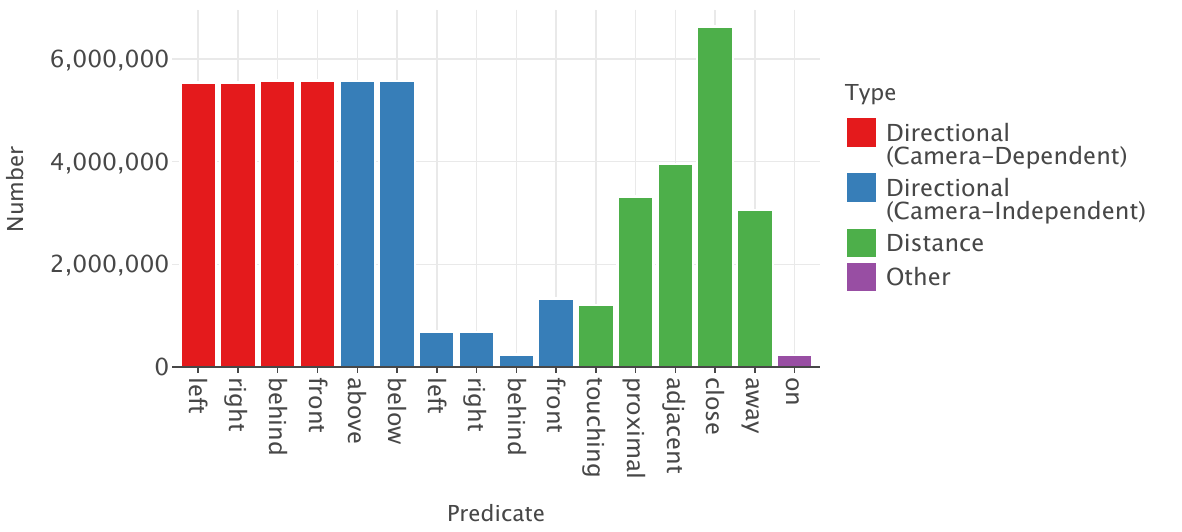}
        \caption{Training Split}
    \end{subfigure}
    \begin{subfigure}[c]{\linewidth}
        \includegraphics[width=\linewidth]{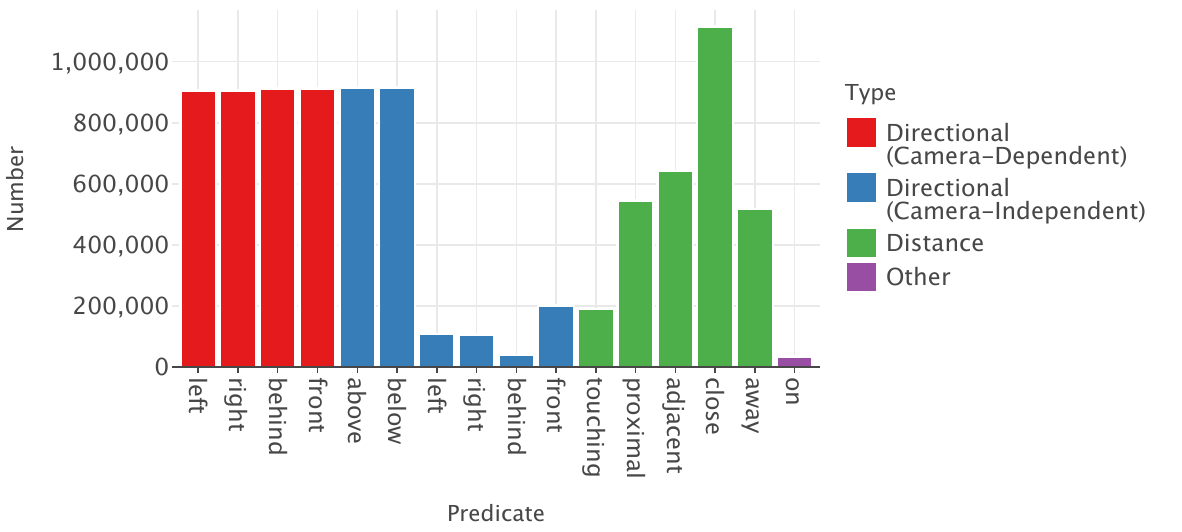}
        \caption{Validation Split}
    \end{subfigure}
    \begin{subfigure}[c]{\linewidth}
        \includegraphics[width=\linewidth]{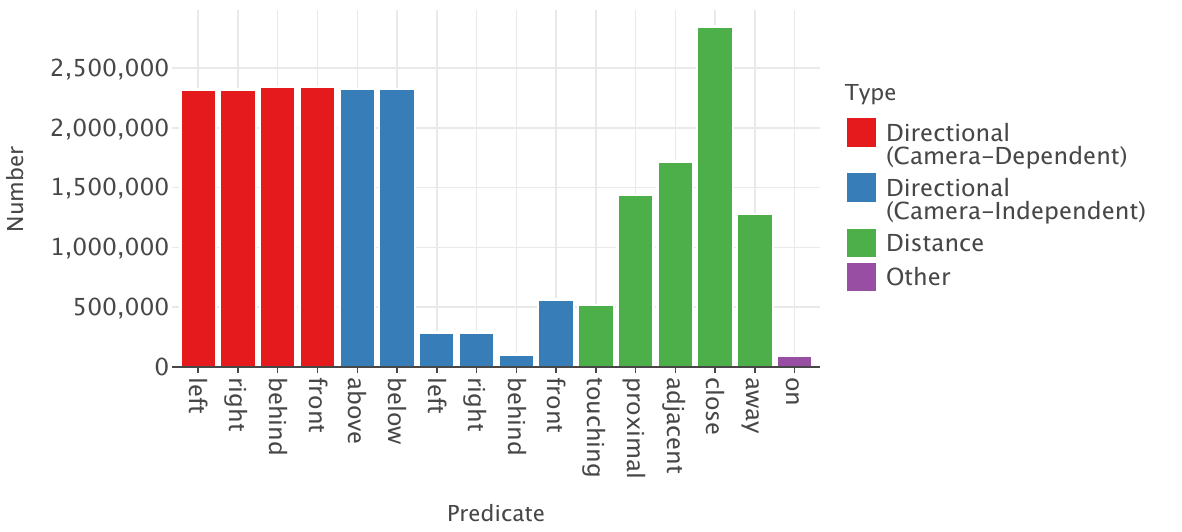}
        \caption{Test Split}
    \end{subfigure}
    \caption{Predicate class distribution in the overall dataset. All distance based relations are stored internally as a single relation. This figure shows a discretized variant for clarity (proximal $\in (0, 0.3]$, adjacent $\in (0.3, 1]$, close $\in (1, 3]$, away $\in (3, \infty)$).}
    \label{fig:pred_distr_splits}
\end{figure}

We define a fixed split of \ourdata{} into training, validation, and test. \cref{fig:pred_distr_splits} shows the distribution of the contained predicate classes among the different data splits.

\end{document}